\pdfoutput=1
\documentclass[11pt]{article}

\usepackage[final]{acl}

\usepackage{times}
\usepackage{latexsym}

\usepackage[T1]{fontenc}
\usepackage[utf8]{inputenc}

\usepackage{microtype}

\usepackage{inconsolata}

\usepackage{graphicx}
\graphicspath{{figures/}}
\usepackage[noend]{algpseudocode}
\usepackage{microtype}
\usepackage{markdown}
\usepackage{booktabs}
\usepackage{multirow}
\usepackage{float}
\usepackage{array}
\usepackage[listings,skins,breakable]{tcolorbox}
\usepackage{amsmath}
\usepackage{subcaption}
\usepackage{algorithm}
\usepackage{amsfonts}
\usepackage{booktabs}
\usepackage[table]{xcolor}
\definecolor{lightgray}{gray}{0.92}
\definecolor{exampleblue}{rgb}{0.88, 0.92, 1.0}
\usepackage{pifont}
\usepackage{tabularx}
\definecolor{goodgreen}{RGB}{0,120,0}
\definecolor{warningyellow}{RGB}{200,140,0}
\definecolor{badred}{RGB}{180,0,0}
\definecolor{softblue}{RGB}{230,240,255}
\definecolor{softgray}{RGB}{245,245,245}
\definecolor{targetgray}{RGB}{80,80,80}

\definecolor{modelblue}{RGB}{0,70,160}
\definecolor{modelpurple}{RGB}{120,0,120}

\def\R{{\mathbb R}}

\title{Linguistically-Controlled Paraphrase Generation}

\author{Mohamed Elgaar \and Hadi Amiri \\
  University of Massachusetts Lowell\\
  \texttt{\{melgaar,hadi\}@cs.uml.edu}}

\begin{document}
\maketitle
\begin{abstract}
Controlled paraphrase generation produces paraphrases that preserve meaning while allowing precise control over linguistic attributes of the output. We introduce \textsc{LingConv}, an encoder-decoder framework that enables fine-grained control over 40 linguistic attributes in English. To improve reliability, we introduce a novel inference-time quality control mechanism that iteratively refines attribute embeddings to generate paraphrases that closely match target attributes without sacrificing semantic fidelity. \textsc{LingConv} reduces attribute error by up to 34\% over existing models, with the quality control mechanism contributing an additional 14\% improvement.\footnote{Our code and an interactive demo~\citep{elgaar-amiri-2025-lingconv} are available at \url{https://github.com/CLU-UML/LingConv}.}
\end{abstract}

\section{Introduction}
Controllable text generation (CTG) aims to produce text with specified linguistic attributes~\citep{ficler-goldberg-2017-controlling, jin-etal-2022-deep}. A sub-task, controlled paraphrase generation (CPG), aims to generate  paraphrases that satisfy desired attributes while preserving meaning. CPG has applications in text simplification~\citep{lee2023prompt, zhang-lapata-2017-sentence, maddela-etal-2021-controllable}, toxicity control~\citep{zheng-etal-2023-invariant}, data augmentation~\citep{iyyer-etal-2018-adversarial}, and creating linguistically challenging data~\citep{perkoff-etal-2023-comparing}. The key challenge is to balance attribute adherence with semantic fidelity.

Prior work in CPG typically controls a small number of attributes, often less than three~\citep{bandel-etal-2022-quality, liu-etal-2023-bolt, yang-etal-2023-tailor}. Large language models (LLMs) such as Llama~\citep{dubey2024llama}, while powerful, struggle with precise and simultaneous control over many attributes via prompting~\cite{dekoninckcontrolled}. In addition, decoding-time methods that use attribute classifiers can be slow and less effective in high-dimensional attribute spaces~\citep{yang-klein-2021-fudge, liu-etal-2023-bolt}, and inference-time quality control is rarely addressed.

CPG can generate linguistically challenging data\footnote{Especially in the current era of NLP, where datasets often contain examples that lack enough linguistic complexity, leading to a plateau in model performance improvements.}~\citep{perkoff-etal-2023-comparing,ashok-kumar-etal-2023-improving,wambsganss-etal-2022-alen}, augment datasets~\citep{iyyer-etal-2018-adversarial,malandrakis-etal-2019-controlled}, and support language simplification~\citep{lin-etal-2021-towards-document-level}. The main challenge is to generate text that preserves meaning and satisfies target attributes. Most prior work focuses on a limited set of attributes. However, broader attribute control increases flexibility for diverse audiences.

\begin{figure}[t]
    \centering
    \includegraphics[width=\linewidth]{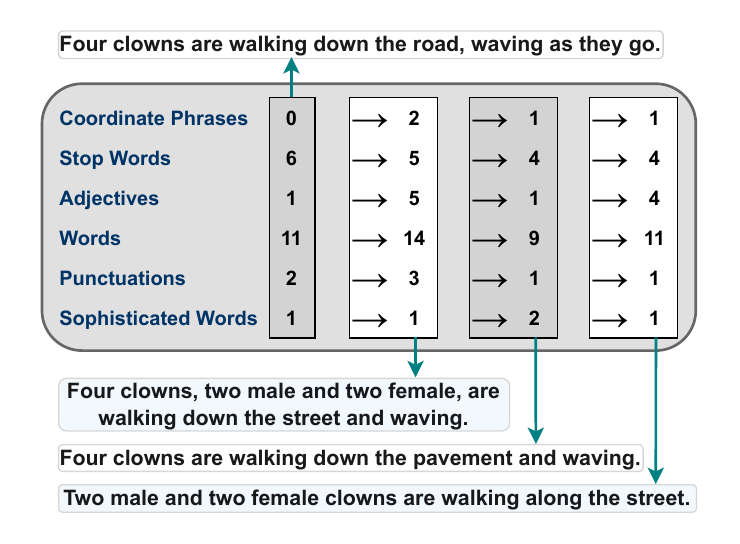}
    \vspace{-25pt}
    \caption{We aim to transform a given sentence into multiple paraphrases, each satisfying distinct linguistic attributes. Our model takes a source sentence and a set of target linguistic attributes and generates a paraphrase optimized to satisfy the target attributes. Here we show three paraphrases with different linguistic attributes generated for the source sentence. Linguistic features identified using the spaCy ``en\_core\_web\_sm'', with stop-word list from \citet{spacy_stopwords}.
    }
    \label{fig:intro}
    \vspace{-10pt}
\end{figure}

We introduce \textsc{LingConv}, a novel encoder-decoder CPG model that offers fine-grained control over 40 linguistic attributes spanning lexical, syntactic, discourse, and semantic aspects (see Appendix~\ref{sec:index_list}). \textsc{LingConv} integrates attribute embeddings directly into the decoder and employs a robust inference-time quality control (QC) mechanism. This QC mechanism iteratively refines outputs using linguistic attribute and semantic consistency classifiers, guided by a line-search algorithm to ensure close alignment with target attributes without sacrificing meaning. Figure~\ref{fig:intro} shows an example of our model's capability.

Our contributions are:
\begin{itemize}
\itemsep0pt
    \item the first system, to our knowledge, for CPG with simultaneous control over 40 fine-grained linguistic attributes;
    \item a novel inference-time quality control mechanism that significantly improves attribute adherence; and 
    \item application to data augmentation, generating attribute-controlled synthetic data to improve downstream task performance.
\end{itemize}
We demonstrate through extensive experiments that \textsc{LingConv} outperforms strong baselines by up to 34\% in attribute control, with the QC mechanism providing a further 14\% improvement.

% Experiments show \textsc{LingConv} outperforms baselines by 34\% in attribute control, with QC providing a further 14\% improvement. We also demonstrate its use in data augmentation, where attribute-controlled synthetic data impacts downstream models differently. Further analysis (Appendix~\ref{app:analysis}) explores which attributes are easy or hard to control.

% Controllable text generation and CPG have advanced, but most prior approaches are limited to a few attributes~\citep{ficler-goldberg-2017-controlling, dathathri2019plug, yang-klein-2021-fudge, bandel-etal-2022-quality, liu-etal-2023-bolt, yang-etal-2023-tailor}, often using discrete tokens or prompts, which lack precision. Decoding-time control with attribute classifiers~\citep{yang-klein-2021-fudge, liu-etal-2023-bolt} is slow and struggles with high-dimensional spaces; quality control at inference is rarely addressed. LLMs like LLaMA~\citep{dubey2024llama} and T5~\citep{2020t5} are strong generators, but prompt-based control is coarse. \textsc{LingConv} addresses these limitations, enabling precise, robust, and efficient paraphrase generation with targeted linguistic properties. See Appendix~\ref{sec:related_work} for a comprehensive review.

CPG has the potential to generate data that challenges existing models from a linguistic perspective,
produce text with varying levels of linguistic complexity for language learners~\citep{okano-etal-2023-generating,
perkoff-etal-2023-comparing,uto-etal-2023-difficulty,
ashok-kumar-etal-2023-improving,wambsganss-etal-2022-alen} or data augmentation~\citep{iyyer-etal-2018-adversarial,malandrakis-etal-2019-controlled}, and make text accessible through language simplification~\citep{lin-etal-2021-towards-document-level}.
The main challenge in CPG is to generate text that preserves the meaning of the source and satisfies the desired linguistic attributes. While existing work has explored this balance, most work has focused on a limited set of attributes, as discussed below.
Accommodating a wider array of linguistic attributes in CPG is crucial because it improves the flexibility and engagement for diverse audiences including language learners.

% Recent works have primarily focused on controlling the generation of a language model (decoder-only model) for a small number of attributes, often less than three.
\textsc{LingConv} is an encoder-decoder CPG model that offers fine-grained control over 40 linguistic attributes spanning lexical, syntactic, discourse, and semantic aspects (see Appendix~\ref{sec:index_list}). It integrates attribute embeddings directly into the decoder and employs a novel inference-time quality control (QC) mechanism that iteratively refines outputs using linguistic attribute and semantic consistency classifiers to ensure close alignment with target attributes without sacrificing meaning.

Extensive experiments demonstrate that \textsc{LingConv} outperforms strong baselines by up to 34\% in attribute control, with the QC mechanism providing a further 14\% improvement. We also show the utility of our approach in data augmentation, where attribute-controlled synthetic data can be tailored to improve downstream task performance. Analysis reveals which linguistic attributes are easier or harder to control and the factors influencing controllability.

\begin{figure*}[htb]
    \centering
    \includegraphics[scale=.9]{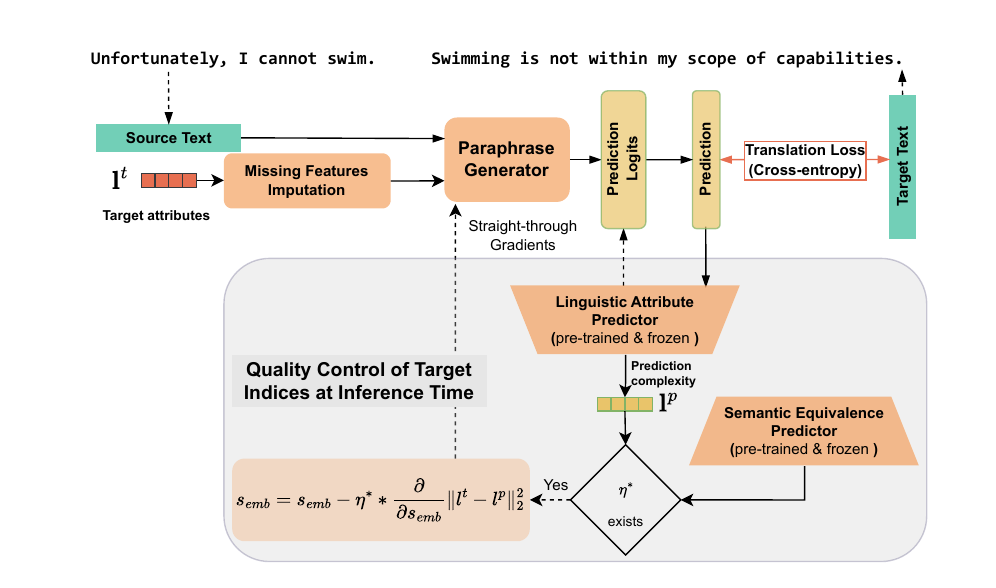}
    \vspace{-15pt}
    \caption{\textsc{LingConv} Architecture:
    The paraphrase generator extends the T5 model by incorporating linguistic attributes into the decoder inputs.
    Linguistic attributes of the source ($\mathbf{l^s}$) and target ($\mathbf{l}^t$) are embedded and fused with the generation using element-wise addition to the decoder inputs.
    In addition, the linguistic attribute predictor estimates attributes of the generated text, which facilitates backpropagation of the linguistic attribute error. During inference, the quality control mechanism iteratively adjusts inputs to guide outputs towards desired attributes. The Semantic Equivalence Predictor (SE) receives as input the source sentence and the candidate generation $\hat{t}$, as in Algorithm~\ref{alg:qc} (line 25), to assess semantic similarity. The model is trained with a dual objective of semantic equivalence and linguistic attribute adherence.}
    \label{fig:model}
    \vspace{-10pt}
\end{figure*}

\section{Related Work}
\label{sec:related_work}
Controllable text generation (CTG) and controlled paraphrase generation (CPG) have seen significant advances, with early works focusing on controlling a small set of attributes such as formality~\citep{ficler-goldberg-2017-controlling, dathathri2019plug, yang-klein-2021-fudge}. Most prior CPG approaches are limited to manipulating up to three attributes simultaneously~\citep{bandel-etal-2022-quality, liu-etal-2023-bolt, yang-etal-2023-tailor}, often relying on discrete control tokens or prompt-based strategies, which can be imprecise and lack fine-grained control. Decoding-time control methods using attribute classifiers~\citep{yang-klein-2021-fudge, liu-etal-2023-bolt} are typically slow and struggle with high-dimensional attribute spaces, and quality control at inference time is rarely addressed. LLMs such as Llama~\citep{dubey2024llama} and T5~\citep{2020t5} demonstrate strong general-purpose generation, but prompt-based control remains coarse and unreliable for fine-grained attribute manipulation.

\citet{colin-gardent-2018-generating} show that including a textual syntactic constraint in paraphrase generation produces syntactically diverse outputs. Other approaches have explored keyword exclusion~\citep{kajiwara-2019-negative}, using discriminator networks to enforce diversity~\citep{qian-etal-2019-exploring}, and following exemplar syntax~\citep{chen-etal-2019-controllable}. FSET~\citep{kazemnejad-etal-2020-paraphrase} improves quality and diversity by retrieving similar paraphrase pairs and applying their edits to the source sentence. variational autoencoders (VAEs) were used to disentangle semantic and syntactic representations to generate diverse paraphrases~\citep{chen-etal-2020-semantically, yang-etal-2021-syntactically, hosking-lapata-2021-factorising}. Models that condition on explicit syntactic sketches via hierarchical discrete codes have also been proposed~\citep{hosking-etal-2022-hierarchical}. GCPG~\citep{yang-etal-2022-gcpg} concatenates conditions to the input to control keywords and syntax. \citet{shi2024action} introduced action-controlled paraphrasing using action tokens, though this does not directly control specific linguistic attributes.

Alternative approaches to CTG include energy-based models that sample from a latent space~\citep{kumar2021controlled, wang2019controllable,gu-etal-2023-controllable,liu-etal-2023-composable}. Gradient-based methods like PPLM~\citep{dathathri2019plug} steer generation using an attribute classifier at inference time but are often slow. FUDGE~\citep{yang-klein-2021-fudge} improves efficiency by re-weighting the next-token probability based on the desired attribute.

Existing work has focused on specific types of control. QCPG~\citep{bandel-etal-2022-quality} controls for three abstract attributes (semantic similarity, syntactic variation, lexical variation), while others focus on keyword presence~\citep{Zeng_Zhang_Xiang_Wang_Ji_2019, liu-etal-2023-bolt}. Syntactically-controlled paraphrase generation has been explored by manipulating abstract meaning representation (AMR) trees~\citep{huang-etal-2023-paraamr}, reordering parse tree segments~\citep{goyal-durrett-2020-neural}, or using constituency parse templates~\citep{iyyer2018adversarial}. Other methods disentangle semantics and syntax by adding sentence parse trees or AMR trees as features~\citep{huang-chang-2021-generating, huang-etal-2022-unsupervised-syntactically}.

In summary, previous works have primarily focused on a narrow set of linguistic attributes and often lack robust quality control mechanisms at inference time. In addition, LLMs are powerful general-purpose generation, but achieving fine-grained, multi-attribute control is still a major challenge. Our work addresses these gaps by introducing a model capable of controlling a large, diverse set of linguistic attributes simultaneously, complemented by a novel inference-time quality control mechanism to ensure both attribute adherence and semantic fidelity.

\section{LingConv}
\subsection{Problem Formulation}
Consider a dataset $\mathcal{D} = \{(s_i, t_i, l_i^t)\}_{i=1}^{N}$, where each triplet contains a source sentence ($s$), a target sentence ($t$), and the target's linguistic attributes ($l^t \in \R^k$). The task is to generate $t$, given $s$ and $l^t$.

\subsection{LingConv Architecture}
\label{sec:model}
\paragraph{Overview}
\textsc{LingConv} is a seq2seq model with three main components (Fig.~\ref{fig:model}): an encoder-decoder paraphrase generator, a linguistic attribute predictor, and a quality control (QC) module. The attribute predictor and a semantic equivalence classifier are pre-trained and used only during inference for QC. Input attribute vector may specify any subset of attributes; missing values are allowed and they are imputed via MICE~\citep{azur2011multiple}. The encoder-decoder integrates attribute embeddings into the generation process, and the QC module iteratively refines outputs to match target attributes. The model is trained to generate paraphrases conditioned on source and target attributes.

\paragraph{Encoder-Decoder}
The encoder-decoder extends T5~\citep{2020t5}, embedding the target attribute vector $l^t$ and adding it to the first decoder input token. Decoder-side injection provides direct, precise control over generation (see Appendix~\ref{app:injection_methods}). This approach balances semantic preservation and attribute control, and allows users to specify only selected attributes, with the remainder imputed from training data patterns.

Specifically, to guide the model toward generating desired outputs, we embed the linguistic attributes $l^t$ into a dense vector representation and integrate them with T5's decoder inputs. While incorporating attributes via input modifications is common in controlled generation, our architecture achieves a balanced trade-off between semantic preservation and attribute control by performing decoder-side injection. 

Our architecture effectively balances semantic preservation and attribute control by injecting attributes at the decoder side, allowing direct influence on token generation through self-attention while maintaining access to the full source context through cross-attention. First-token injection strikes an optimal balance between providing a strong control signal and minimizing disruption to the pre-trained model's capabilities.

To address practical concerns regarding the specification of all linguistic attributes, our approach can utilize variable imputation. This allows users to specify only the variables of interest, while the model fills in the rest based on learned patterns from the training data.

Specifically, in order to effectively guide the model toward generating desired outputs, we propose to embed the linguistic attributes $l^t$ 
% indices 
into a dense vector representation and integrate it with T5's {\em decoder} inputs.
% \footnote{We also experimented with adding linguistic embeddings to all tokens of the decoder input, concatenating to the decoder inputs (equivalent to prompt tuning), concatenation/addition to encoder inputs, concatenating/adding to encoder outputs, and fusing to encoder outputs using a linear layer. In general, decoder injections were better than encoder injections. Decoder first-token-addition was the best-performing overall.}
To achieve this goal, we add the embedding of the target linguistic vector $\mathbf{l}^t$ to the first token of the decoder inputs, which corresponds to the beginning of sentence token {\tt <bos>}: 
\begin{equation}
\label{eq:decoder-add}
    Y'(l^t) = 
    \begin{cases}
        Y_i \otimes \mathrm{LE}(l^t) & \text{if } i = 0\\
        Y_i & \text{otherwise,}
    \end{cases}
\end{equation}
where 
% $D$ is the decoder module, 
$Y$ is the decoder input embedding, $\mathrm{LE}$ is the linguistic attribute embedding layer, $\otimes$ indicates the element-wise addition operation, and $Y'$ is the updated decoder inputs.
$\mathrm{LE}$ is a fully connected layer from $\R^k$ to $\R^d$, where k is the number of linguistic attributes and $d$ is the dimension of text input embeddings.
The input attributes are standardized to a mean of 0 and a variance of 1 prior to embedding.
% The purposes of $\mathrm{LE}$ is to compress the $k$ linguistic indices into a dense vector representation and to project the linguistic indices into $\R^d$, such that they can be integrated with token embeddings.

\paragraph{Objective}
We train our model using cross entropy loss (\ref{eq:ce}):
\begin{equation}\label{eq:ce}
    \ell_{CE}(s_i, t_i) = \sum_{j=0}^{len(y)-1}-\log{p(y_i^{(j)}|x_i, y^{<j})},
\end{equation}
where $p(y_i^{(j)}|x_i, y^{<j})$ is the probability of the model predicting the $j$-th token in the target sequence given the source sequence $x_i$ and the previous tokens $y^{<j}$ in the target sequence; this loss translates the source sentence to a semantically equivalent sentence as induced by our choice of training data (only paraphrase examples). 
At test time, the model takes a source sentence, the linguistic attributes of the source sentence, and the desired linguistic attributes; and generates an output using auto-regressive greedy decoding.

\paragraph{Linguistic Attribute Predictor}(LP) estimates the linguistic attributes of a given generation. This component is independently pre-trained and frozen. It allows for differentiable computation of linguistic attributes and thus backpropagation of the error. 
While existing linguistic tools can extract attributes, they are not differentiable and would require reinforcement learning approaches for optimization.
Moreover, it helps us avoid the computationally intensive task of calculating 40 linguistic attributes for each generated text within the training process.
We implement the linguistic predictor ($\mathrm{LP}$) using a T5 encoder followed by a projection layer, and it is pre-trained by minimizing the mean squared error of the predicted linguistic attributes of each text ($\mathrm{LP}(x)=l^p$ in Figure~\ref{fig:model}) from its gold attributes ($l^x$) as follows:
\begin{equation}
\label{eq:disc}
    \ell_{disc}(x) = \| \mathrm{LP}(x) - l^x \|_2^2.
\end{equation}
% The linguistic predictor and semantic equivalence model are both fully pre-trained for 400k steps before training the main model.
% During training, our controlled paraphrase generation model estimates paraphrases for input source sentences. These estimations are fed into the linguistic attribute predictor to minimize the error between $\mathbf{l}^t$ and $\mathbf{l}^p$, where $\mathbf{l}^p$ are the linguistic attributes of the prediction.
It is not possible to backpropagate the loss through a discrete prediction resulting from an {\em argmax} operation. Therefore, we apply Straight-through Gradient Estimation~\citep{bengio2013estimating} to the linguistic attribute predictor, so the gradient is propagated to the prediction logits through the multiplication of the prediction logits and the regressor's token embedding matrix, further described in Appendix~\ref{app:ste}.
Additionally, the LP enables baseline methods like BOLT~\citep{liu-etal-2023-bolt} and FUDGE~\citep{yang-klein-2021-fudge} that require differentiable attribute scoring for decoding-time control.

During inference, this pre-trained LP is used within the QC mechanism to evaluate how well the generated text $\hat{t}$ matches the target attribute vector $l^t$. Specifically, LP computes the attribute error $\|\mathrm{LP}(\hat{t}) - l^t\|_2^2$, which is then used to guide the iterative refinement of the generated output, as detailed in Algorithm~\ref{alg:qc}. 

% \footnote{We also experiment with the feedback of the semantic equivalence module, but we find no improvement in performance, so we only use the linguistic discriminator feedback.}

\paragraph{Imputation of Missing Values} Manually specifying all 40 linguistic attributes for a target paraphrase is impractical and prone to error, as users may not know desirable values for every attribute or may specify inconsistent combinations. To address this real-world challenge, we employ the Multiple Imputation by Chained Equations (MICE) algorithm~\citep{azur2011multiple}. This allows users to provide values for only a small subset of attributes they wish to control. The model then imputes the remaining values by leveraging statistical relationships learned from the training data. This feature significantly improves usability and reduces the risk of misconfiguration. See Appendix~\ref{app:imputation_details} for details.

\paragraph{Semantic Equivalence Classifier}(SE) quantifies semantic equivalence of a pair of sentences, and is used in the quality control algorithm.
The SE module receives as input the source sentence and the candidate generation $\hat{t}$ to compute semantic equivalence.
We implement SE using a T5 encoder followed by a projection layer. This design ensures architectural compatibility and efficient integration with our T5-based LingConv generation model, and allows us to pre-train SE using a contrastive loss function specifically tailored to our paraphrase data. Notably, the contrastive loss described below was used exclusively for pre-training SE; it was not explored during the main LingConv model training. During inference, SE serves solely as a fixed, pre-trained module for semantic equivalence assessment, without further updates or integration of the contrastive loss into the primary model's training objective. SE is pre-trained by minimizing the following contrastive loss:
\begin{equation}
\label{eq:sem}
    \ell_{sem}(s, t) = - \log{\frac{\text{SE}(s,t)}{\sum\limits_{t' \in \mathcal{N}(s)}\text{SE}(s,t')}},
\end{equation}
where $\mathcal{N}(s)$ is the set of negative paraphrases of $s$. The loss maximizes the probability of valid paraphrases $(s,t)$ and minimizes the probability of invalid paraphrases $(s,t')$. For a mini-batch of size $m$, $m-1$ samples are used as negative paraphrases for the remaining sample.

\begin{algorithm}[h!]\small
\caption{\textbf{Quality Control}\\
\small{This algorithm optimizes the alignment of generated text with target linguistic attributes while preserving semantic equivalence to the source. The quality control loop adjusts the text embeddings iteratively using a gradient-based method combined with a line search to minimize attribute errors. The process continues until a satisfactory generation is found or the algorithm exhausts its search.}}
\begin{algorithmic}[1]
\Require model $M$, linguistic predictor $LP$, semantic classifier $SE$, input $s$, target attributes $l^t$, base step size $\eta_0$, step size scaling factor $\gamma$, semantic equivalence threshold $\tau$, patience $k$
\Procedure{Quality\_Control}{$s, l^t$}
    \State $\Theta \gets Emb(s)$ \Comment{Initialize embeddings from the source text}
\While{True}
    \State $\hat{t} \gets M(\Theta, l^t)$ \Comment{Generate text with current embeddings}
    \State $l_{\text{current}} \gets \|LP(\hat{t}) - l^t\|_2^2$ \Comment{Compute attribute error}
    \State $g \gets \nabla_\Theta l_0$ \Comment{Compute gradient w.r.t. embeddings}
    \State $\Theta \gets$ \Call{Adaptive\_Step\_Search}{$\Theta, l_0$}
    \If{$\Theta = null$}
        \State \textbf{break} \Comment{Terminate if no improvement is found}
    \EndIf
\EndWhile
\State \textbf{return} $\hat{t}$
\EndProcedure
\Procedure{Adaptive\_Step\_Search}{$\Theta, l_0$}
    \State $\eta \gets \eta_0$ \Comment{Initialize step size}
    \State $\text{patience} \gets k$ \Comment{Initialize patience counter}
    
    \While{patience $> 0$}
        \State $\sigma_{\text{sem}} \gets SE(s, \hat{t}')$ \Comment{Check semantic equivalence}
        
        \If{$l' < l_0$ and $\sigma_{\text{sem}} \geq \tau$}
            \State \textbf{return} $\Theta'$ \Comment{Accept and return the new embeddings}
        \Else
            \State $\eta \gets \eta * \gamma$ \Comment{Reduce step size}
            \State $\text{patience} \gets \text{patience} - 1$ \Comment{Decrease patience}
        \EndIf
    \EndWhile
    
    \While{patience $> 0$}
        \State $\Theta' \gets \Theta - \eta * g$ \Comment{Update embeddings}
        \State $\hat{t}' \gets M(\Theta', l^t)$ \Comment{Generate text}
        \State $l' \gets \|LP(\hat{t}') - l^t\|_2^2$ \Comment{Compute new attribute error}
        \State $\sigma_{\text{sem}} \gets SE(s, \hat{t}')$ \Comment{Check semantic equivalence}
        
        \If{$l' < l_0$ and $\sigma_{\text{sem}} \geq \tau$}
            \State \textbf{return} $\Theta'$ \Comment{Accept and return the new embeddings}
        \Else
            \State $\eta \gets \eta * \gamma$ \Comment{Reduce step size}
            \State $\text{patience} \gets \text{patience} - 1$ \Comment{Decrease patience}
        \EndIf
    \EndWhile
    \State \textbf{return} $null$ \Comment{Return null if no improvement}
\EndProcedure
\end{algorithmic}
\label{alg:qc}
\end{algorithm}

\paragraph{Quality Control}
To ensure high-quality outputs, we propose a quality control mechanism to use at inference time. Achieving precise control over multiple linguistic attributes while maintaining text quality presents significant challenges in controlled text generation. Our approach employs an adaptive, gradient-based iterative refinement process~\citep{padmakumar2023extrapolative} that dynamically adjusts the model's input embeddings to steer outputs toward the target attributes. To ensure both attribute control and semantic fidelity, we use a line search algorithm~\citep{armijo1966minimization,boyd2004convex} to select the optimal update strength at each step. This mechanism enables robust, fine-grained control over linguistic properties during inference. For a detailed description of the algorithm, see Appendix~\ref{app:qc_details}.

\section{Experiments}
We evaluate \textsc{LingConv} on MRPC, STS-B, and QQP, using BERTScore and mean squared error (MSE) of attribute adherence (see Appendix~\ref{sec:data}).

\begin{table*}[t]
    \centering
    \small
    \begin{tabular}{l llll|llll}
    \toprule
     & \multicolumn{4}{c}{}& \multicolumn{4}{c}{\textbf{Novel Target Challenge}}\\
         \textbf{Model} &  \textbf{BERTS}$\uparrow$& \textbf{MSE}($\pmb{l^t}$)$\downarrow$& \textbf{MSE}($\pmb{l^s}$)$\uparrow$& \textbf{Overall}$\uparrow$&\textbf{BERTS}$^F\uparrow$& \textbf{MSE}($\pmb{l^{t}}$)$\downarrow$&\textbf{MSE}($\pmb{l^s}$)$\uparrow$&\textbf{Overall}$\uparrow$\\
         \midrule
\textbf{Ref} & 100.0 & 0.00 & 0.96 & 0.85 & 94.4 & 9.82 & 0.96 & 0.19 \\
\textbf{Copy} & 94.4 & 0.96 & 0.00 & 0.32 & 100.0 & 9.86 & 0.00 & 0.33 \\
\textbf{T5-FT} & 94.24 & 0.96 {\tiny$\pm$ 0.03} & 0.51 {\tiny$\pm$ 0.04} & 0.48 {\tiny$\pm$ 0.01} & 96.65 & 9.00 {\tiny$\pm$ 0.78} & 0.68 {\tiny$\pm$ 0.02} & 0.32 {\tiny$\pm$ 0.03} \\
\midrule
\textbf{Llama} & 91.03 & 2.24 {\tiny$\pm$ 0.08} & \textbf{1.86} {\tiny$\pm$ 0.07} & 0.38 {\tiny$\pm$ 0.01} & \textbf{92.77} & 8.71 {\tiny$\pm$ 0.49} & 2.47 {\tiny$\pm$ 0.26} & 0.27 {\tiny$\pm$ 0.03} \\
\textbf{BOLT} & 90.64 & 1.12 {\tiny$\pm$ 0.03} & 1.10 {\tiny$\pm$ 0.05} & 0.44 {\tiny$\pm$ 0.01} & 90.38 & 7.34 {\tiny$\pm$ 0.66} & 1.84 {\tiny$\pm$ 0.06} & 0.29 {\tiny$\pm$ 0.04} \\
\textbf{FUDGE} & 92.01 & 0.85 {\tiny$\pm$ 0.01} & 1.05 {\tiny$\pm$ 0.05} & 0.48 {\tiny$\pm$ 0.01} & 92.53 & 6.94 {\tiny$\pm$ 0.78} & 2.92 {\tiny$\pm$ 0.57} & 0.32 {\tiny$\pm$ 0.05} \\
\textbf{QCPG} & \textbf{95.36} & 0.63 {\tiny$\pm$ 0.02} & 0.82 {\tiny$\pm$ 0.05} & 0.54 {\tiny$\pm$ 0.01} & 91.36 & 5.54 {\tiny$\pm$ 0.55} & 3.14 {\tiny$\pm$ 0.06} & 0.36 {\tiny$\pm$ 0.04} \\
\textbf{Lingconv} & 95.15 & 0.62 {\tiny$\pm$ 0.04} & 0.80 {\tiny$\pm$ 0.06} & 0.55 {\tiny$\pm$ 0.01} & 92.04 & 3.92 {\tiny$\pm$ 0.32} & 4.20 {\tiny$\pm$ 0.30} & 0.41 {\tiny$\pm$ 0.03} \\
\textbf{\hspace{1mm} +QC} & 95.17 & \textbf{0.56} {\tiny$\pm$ 0.04} & 0.77 {\tiny$\pm$ 0.05} & \textbf{0.56} {\tiny$\pm$ 0.01} & 91.54 & \textbf{3.07} {\tiny$\pm$ 0.29} & \textbf{5.92} {\tiny$\pm$ 0.37} & \textbf{0.44} {\tiny$\pm$ 0.04} \\
\bottomrule
    \end{tabular}
    \caption{
    Mean squared error (MSE) values
    reflect how close the linguistic attributes of the generated paraphrase are to the target (MSE($\pmb{l^t}$)$\downarrow$) or source (MSE($\pmb{l^s}$)$\uparrow$). Lower MSE($\pmb{l^t}$) indicates better attribute control; higher MSE($\pmb{l^s}$) indicates greater deviation from the source. Results are averaged over three seeds; standard error is shown for all metrics except BERTScore, where it is always less than 0.01.
}
    \label{tab:results}
    \vspace{-10pt}
\end{table*}

\begin{table}[t]
    \centering
    \small
\begin{tabular}{lllll}
\textbf{Model} & \textbf{Lexical} & \textbf{Syntactic} & \textbf{Discourse}  & \textbf{Macro-} \\
& & & & $\textbf{MSE} (\pmb{l^t})$ \\
\toprule
\textbf{Ling-disc} & 0.08 & 0.14& 0.50 & 0.24 \\
\bottomrule
\end{tabular}
    \caption{Pre-training test loss of the linguistic discriminator.}
    \label{tab:ling-disc-loss}
    % \vspace{-10pt}
\end{table}

\subsection{Experimental Setup}
For each source and target sentence in our dataset, we extract the 40 linguistic attributes (listed in Appendix~\ref{sec:index_list}) from existing linguistic toolkits~\citep{Lu_2010, lu2012relationship,lee-lee-2023-lftk,elgaar-amiri-2023-ling}. The attributes include lexical, syntactic, semantic, and discourse attributes, which capture a comprehensive spectrum of linguistic structures. We use {\texttt flan-t5-base}~\citep{chung2024scaling} as a base model, and re-implement all baselines to use the same base model for fairness.
We use greedy decoding for all models.

Furthermore, we compare against: Copy (input as output), Reference (gold paraphrase), T5-FT (fine-tuned T5), FUDGE~\citep{yang-klein-2021-fudge}, QCPG~\citep{bandel-etal-2022-quality}, BOLT~\citep{liu-etal-2023-bolt}, and Llama~\citep{dubey2024llama}. See Appendix~\ref{app:baselines} for a detailed description of each baseline.

\subsection{Evaluation}
\label{sec:evaluation}
Our evaluation is designed to assess both semantic fidelity and fine-grained linguistic control in our controlled paraphrase generation system. In the standard evaluation setting, where target reference paraphrases are available, we adopt \textbf{BERTScore}~\citep{Zhang*2020BERTScore:} to measure semantic similarity between the generated paraphrase and its corresponding reference. BERTScore leverages contextualized embeddings to capture deep semantic correspondences that go beyond surface-level n-gram overlap, making it particularly effective in scenarios with substantial linguistic reformulation.

To quantify the model's ability to adhere to target linguistic attributes, we measure the mean squared error between the generated paraphrase's linguistic attributes and the target attributes, denoted as \textbf{MSE($\pmb{l^t}$)}. Lower values of MSE($\pmb{l^t}$) indicate that the generated paraphrase closely follows the desired attribute controls. Furthermore, we compute \textbf{MSE($\pmb{l^s}$)} to assess the divergence of the paraphrase from its source text, ensuring that the output not only preserves the intended semantic content but also exhibits the required linguistic modifications.

A lower MSE($\pmb{l^t}$) indicates better attribute control, a higher MSE($\pmb{l^s}$) is actually desirable in many cases, as it reflects the model's ability to produce significant linguistic transformations from the source. This is particularly important in the Novel Target Challenge, where successful models must demonstrate the capacity to significantly restructure inputs according to target attributes that differ substantially from the source text's attributes. A model that simply copies the source (or makes minimal changes) would have a low MSE($\pmb{l^s}$), indicating insufficient attribute transformation.

To provide a concise summary of performance across these dimensions, we define an \textbf{Overall} score computed as the average of three normalized metrics, each scaled to lie between 0 and 1: BERTScore, the normalized MSE($\pmb{l^s}$), and $(1-\text{normalized MSE}(\pmb{l^t}))$. This Overall score captures our dual objectives of preserving semantic fidelity and effective attribute control.

In addition to standard evaluation, we introduce the \textit{Novel Target Challenge}, a more demanding setting in which models generate paraphrases based on target linguistic attributes derived from an ``irrelevant'' sentence relative to the source. An ``irrelevant'' sentence is one randomly sampled from the test set, with no guaranteed semantic or topic relation to the source. This creates a robust test of a model's ability to generate diverse paraphrases, independent of the source's linguistic structure. Since no gold reference is available in this scenario, we employ a reference-free variant of BERTScore~\citep{shen-etal-2022-evaluation}, denoted by \textbf{BERTScore$^F$}. Reference-free BERTScore computes semantic similarity directly with respect to the source text instead of a gold standard reference, thereby providing a robust assessment when the target attributes are decoupled from conventional reference paraphrases. This is crucial for testing model adaptability in real-world applications where specified target attributes may be entirely novel relative to the source.

Alternative metrics such as iBLEU~\citep{liu-etal-2020-unsupervised,niu-etal-2021-unsupervised} have been proposed for paraphrase evaluation to balance semantic similarity with lexical diversity by penalizing excessive overlap with the source. However, these metrics focus largely on surface-level comparisons. In contrast, our evaluation framework, which combines BERTScore (or BERTScore$^F$ in the novel target setting) with MSE metrics for target and source linguistic attributes, directly quantifies both semantic preservation and the degree to which controlled linguistic attributes are followed, regardless of whether they are similar or different from those of the source. This approach is more aligned with linguistically controlled paraphrase generation.

Detailed analysis of linguistic attribute control and a full description of our paraphrase generation for data augmentation are presented in Appendix~\ref{app:analysis} and Appendix~\ref{app:augmentation}.

\section{Results} \label{sec:lingcontrol}

Table~\ref{tab:results} shows the results obtained by all models across evaluation metrics. 

\subsection{Attribute Control vs. Semantic Fidelity}

Our first observation is that \textsc{LingConv} generates paraphrases that align more precisely with the desired linguistic attributes, as demonstrated by its lower MSE($l^t$) compared to other competing baselines. This result can be attributed to directly integrating linguistic attributes with the decoder input through element-wise addition and the linguistic attribute predictor which effectively guides the decoder to generate paraphrases that adhere to the target linguistic attributes.
% from the superposition of token and attribute embeddings. 
QCPG shows similar MSE($l^t$) performance but it employs a more indirect method for incorporating target attributes---by prefixing the input sequence with special discrete tokens. While effective, this approach may not provide the same level of precision in guiding the generation process. The discrete token prefixes could potentially introduce ambiguity or weaken the direct influence of linguistic attributes on the generated text.
%may not strongly encode and enforce the desired linguistic attributes. 

Second, we observe that \textsc{LingConv} performs well in balancing attribute control, and semantic similarity of output, as shown by the overall score. The balance between attribute control and paraphrase faithfulness is a crucial aspect of high-quality controlled paraphrase generation.
Specifically, within the novel target case \textsc{LingConv} achieves a substantial 34\% decrease in attribute error compared to the best-performing baseline while maintaining comparable semantic consistency as the gold reference paraphrases. 
Furthermore, in the novel target challenge, our quality control approach provides a significant reduction in MSE($\pmb{l^t}$) of the linguistic attributes with minimal reduction in BERTScore, providing a 14\% further decrease in attribute error.

\subsection{Trade-offs in the Novel Target Challenge}

In the Novel Target Challenge, \textsc{LingConv} sometimes achieves a slightly lower BERTScore compared to baselines. This can be explained as a trade-off inherent to the task: \textsc{LingConv} is designed to prioritize adherence to the specified (and often difficult or even conflicting) novel attribute targets, as evidenced by its much lower MSE($l^t$). Achieving these attribute targets may require the model to make more substantial changes to the source sentence, which can result in greater semantic deviation from the original text. In contrast, models that are less effective at precise attribute control (and thus have higher MSE($l^t$)) tend to produce outputs that remain closer to the source, thereby achieving a higher BERTScore, at the cost of failing to achieve the requested attribute modifications.

Third, the novel target case shows LingConv scores a significant increase in MSE($l^s$) compared to the baseline models, with a difference of 2.95 points. 
The low value of MSE($l^s$) indicates that baseline CPG methods are biased by the linguistic structure of the source sentence, and do not deviate far from it, while LingConv can restructure the input sentence to achieve the desired control attributes. Detailed per-dataset results are available in Appendix Table~\ref{tab:dataset_results}, showing that our approach consistently outperforms baselines across all three datasets.

\subsection{Analysis of Baseline Methods}

In addition, we find that BOLT has a limited capacity on fine-grained attribute control. In the novel targets case, BOLT achieves a 24\% drop in error compared to T5-FT, which indicates that it moves in the correct direction. However, it still has a high MSE compared to other CPG methods, indicating that it struggles to control many attributes at once.
On the other hand, FUDGE, with a high enough $\lambda_{FUDGE}$, has a guarantee to reduce the attribute error compared to T5-FT, because it samples the next token with the joint maximum LLM likelihood and minimum attribute error. However, FUDGE has difficulty performing linguistic controls because it relies on long-scale dependencies of the text, where the generation needs to be based on sentence-level decisions rather than token-level.

\begin{table*}[htb]
    \centering
    \small
\begin{tabular}{lcccccc}
 & \multicolumn{2}{c}{\textbf{CoLA (Matthew's Corr.)}} & \multicolumn{2}{c}{\textbf{RTE (Acc.)}} & \multicolumn{2}{c}{\textbf{SST-2 (Acc.)}} \\
\cmidrule(lr){2-3} \cmidrule(lr){4-5} \cmidrule(lr){6-7}
\textbf{Augmentation} & \textbf{Limited Data} & \textbf{Full Data} & \textbf{Limited Data} & \textbf{Full Data} & \textbf{Limited Data} & \textbf{Full Data} \\
\midrule
\textbf{No Aug.} & 53.8 {\scriptsize $\pm$ 0.4} & 60.6 {\scriptsize $\pm$ 1.0}& 68.4\% {\scriptsize $\pm$ 1.5}& 74.2\% {\scriptsize $\pm$ 1.5}& 91.3\% {\scriptsize $\pm$ 0.1}& 92.4\% {\scriptsize $\pm$ 0.3}\\
\textbf{Ineffective Aug.} & 52.5 {\scriptsize $\pm$ 0.8}& 58.4 {\scriptsize $\pm$ 1.1}& 66.1\% {\scriptsize $\pm$ 2.8}& 71.7\% {\scriptsize $\pm$ 2.6}& 91.0\% {\scriptsize $\pm$ 0.3}& 91.7\% {\scriptsize $\pm$ 0.1}\\
\textbf{Effective Aug.} & \textbf{54.8} {\scriptsize $\pm$ 0.6}& \textbf{60.8} {\scriptsize $\pm$ 1.1}& \textbf{71.2\%} {\scriptsize $\pm$ 1.3}& \textbf{76.0\%} {\scriptsize $\pm$ 0.8}& \textbf{92.2\%} {\scriptsize $\pm$ 0.3}& \textbf{93.0\%} {\scriptsize $\pm$ 0.4}\\
\bottomrule
\end{tabular}
\caption{Performance on GLUE tasks with No, Effective and Ineffective augmentation. Effective and ineffective augmentations differ in the set of target linguistic attributes used to generate them.
    % The dataset contains an intrinsic label of difficulty that is used to evaluate the performance.
    }
\label{tab:downstream}
\end{table*}

\subsection{Comparison with Large Language Models}

We observe that LLama, although able to generate %fluent and 
semantically similar paraphrases, has difficulty following instructions for attribute controls. In the standard case, this is evident by the MSE($l^t$) higher than T5-FT, and in the novel target case we see that LLama slightly follows the attribute controls, achieving a poor error comparable to that of T5-FT.

Our model achieves a 34\% error reduction over LLama in attribute control. While large models like LLama-70B excel at general-purpose generation, our results show they struggle with precise attribute control (MSE($l^t$) of 8.90 vs LingConv's 3.69 in novel target scenarios). This highlights a fundamental limitation of prompt-based approaches—even with detailed instructions, LLMs lack the specialized architecture and optimization procedures needed for fine-grained attribute matching. The quality control mechanism provides an additional 14\% error reduction, demonstrating the value of having full access to model gradients and intermediate states, which is not possible with black-box LLM APIs. These results suggest that specialized fine-tuned models remain state-of-the-art for narrow, well-defined tasks requiring precise control and guarantees. For a qualitative comparison of model outputs with attribute-level analysis, see Appendix~\ref{sec:qualitative}.

\subsection{Application in Data Augmentation}
\label{app:augmentation}
We demonstrate the utility of \textsc{LingConv} for data augmentation on three GLUE tasks: CoLA, SST-2, and RTE. By generating paraphrases of training samples with specific linguistic properties, we show that the effectiveness of augmentation is highly dependent on the attributes of the synthetic data. We created ``Effective'' and ``Ineffective'' sets of augmented data by sampling target attributes to either increase or decrease the prevalence of certain linguistic features. Our results (Table~\ref{tab:downstream}) show that ``Effective'' augmentation yields statistically significant performance improvements on downstream tasks, while ``Ineffective'' augmentation can harm performance. This highlights the importance of controlled generation for creating high-quality, targeted training data.

The number of training and test samples for CoLA, SST-2, and RTE are 8.5k and 1k, 67k and 1.8k, and 2.5k and 3k, respectively.  
Data augmentation is generally more effective for smaller datasets~\citep{okimura-etal-2022-impact,louvan-magnini-2020-simple}. 
% Therefore, we use two versions of each dataset, Full and Limited. The Limited version contains a reduced percentage of the original training data (10\% for CoLA and SST-2, and 40\% of RTE due to its smaller size).
Therefore, we use Full and Limited versions of each dataset, with Limited containing reduced training data (10\% for CoLA and SST-2, and 40\% of RTE due to its smaller size).
We use \textsc{LingConv} to generate paraphrases of the training samples, which are added back to the training set with labels matching the original samples.
We create two sets of target attribute vectors by non-uniform sampling from the original data's linguistic attribute vectors ($\mathcal{T}$).
Biased sampling aims to produce increased or decreased prevalence of particular attributes in the generated paraphrases for augmentation, compared to the original data.
This approach allows us to identify which attribute values result in ``Effective'' vs. ``Ineffective'' augmentation based on task performance post-augmentation, compared to no augmentation.
For example, we may sample data such that
$p(l^t\mathbin{:} l^t \in \mathcal{T}) = 0.9 \text{ if } l^t_{\text{\scriptsize[TTR]}} > 0.8$ and
$p(l^t\mathbin{:} l^t \in \mathcal{T}) = 0.1$ otherwise, which results in substantial 
prevalence of high TTR values in the augmented samples.

We run experiments with DeBERTa$_\text{base}$~\citep{he2021deberta}, using the same parameters as their GLUE benchmark experiments. Each experiment is run with six random seeds, and we report the mean and standard error.
We identify ``Effective'' and ``Ineffective'' sets by first evaluating 20 randomly sampled sets. From these, we select two sets: one that shows a statistically significant performance increase and one that shows a significant decrease compared to no augmentation. We then compare the attribute distributions of these two sets to identify which attributes differ significantly.
Results in Table~\ref{tab:downstream} confirms that the distribution of the target attributes influence the effectiveness of data augmentation.

We find that on RTE (Limited), for effective augmentation, target attributes should have a significantly higher prevalence of shorter sentences, while ineffective augmentation produces more medium-length sentences. The Mann–Whitney U test confirms significant differences with p-value $< 0.05$ in the attribute distributions between effective and ineffective sets across all our six datasets. Details are provided in Appendix~\ref{app:aug-data}.

% \textsc{LingConv} can be applied to any new text to produce data of diverse or desired complexity. Therefore, there is no limitation to the use of this model for CPG.

\section{Conclusion}
We present a model for controllable text generation, offering control over 40 linguistic attributes and an effective mechanism for quality control at inference time, yielding a 12\% improvement in output quality.
% We perform extensive experiments
% that show that our method significantly outperform competing baselines in getting in generating text in  target attributes.
% across different settings and evaluation criteria. 
% We introduce a new challenging experimental setup termed ``Novel Target Challenge,'' where competing models should generate paraphrases that adhere to target linguistic attributes associated with an ``irrelevant'' sentence to the source.
We introduce the ``Novel Target Challenge'', where models generate paraphrases based on attributes from an ``irrelevant'' sentence.
The setting evaluates models' adaptability to novel attributes and acts as a robust test for controlled paraphrase generation models.
In addition, we evaluate the model on the downstream application of generating synthetic data for augmentation. Our model generates paraphrases that boost performance and can be used to mitigate dataset biases.
% Future work can investigate alternative evaluation setups, such as using triplets of <source, target, irrelevant> to analyze how a distractor input impacts generation.
Future work can investigate mechanisms to handle contradictory or noisy attribute specifications to enable the model to resolve conflicts and prioritize constraints, and 
extend \textsc{LingConv} beyond English to multilingual and low-resource settings, where new attribute extractors and cross-lingual transfer will be needed.
% where the target attributes are near to the source attributes and when they are "far". We analyze the performance of different models in controlling different types of linguistic indices and provide explanations for the different results for \textit{near} and \textit{far} target attributes and for the gains achieved by the quality control's iterative refinement of text.

% Overall, our model provide significant advancements to controlled text generation with application in data augmentation
% , offering researchers and practitioners control over linguistic attributes in text generation tasks. 
% We believe that our work holds promise for addressing various challenges in text generation and understanding.

% \clearpage
\section*{Limitations}

% Our work has some limitations. % While our analysis sheds light on model behavior and attribute control, it also exposes limitations in current approaches. 
Our approach requires the availability of linguistic attributes, which, although available for the English language, may not be available for all languages. 
Certain linguistic attributes may require more sophisticated control mechanisms. 
% Additionally, our analysis primarily focuses on MSE values, and incorporating more linguistic evaluation metrics could provide a more comprehensive understanding of model performance.
% Although we conduct our study using the English language, the availability requirement may not be fulfilled for other languages.
The direct injection of embedded linguistic attributes into the decoder input in \textsc{LingConv}, although effective, has weaknesses. Specifically, we find it to be sensitive to outlier linguistic targets. If the linguistic target contains extreme values, we find that 
% QCPG still generates reasonable outputs (although not adhering to the extreme target), while the output of 
the model degenerates into non-grammatical and repetitive text. 
% Our choice of injection only to the first decoder token greatly increases the robustness of {LingConv}, however, it is not perfectly robust.
% Furthermore, our approach requires all linguistic attributes as input, when it may only be desired to modify a only a few indices of the source and keep the rest unchanged. 
% work should accommodate the conversion of single attributes more easily and effectively. Moreover, our model currently requires fine-tuning all parameters of the model. 
% Future work may explore parameter-efficient fine-tuning (PEFT) methods~\citep{hu2021lora,houlsby2019parameter}.

In addition, while \texttt{flan-t5-base}~\citep{chung2024scaling} was likely exposed to the training sets of our evaluation tasks during its pre-training, we conduct all evaluations strictly on held-out test sets. We believe any potential data contamination is mitigated because our core challenge is fine-grained attribute control, and our evaluation uses novel attribute combinations and reference-free metrics in the challenging scenarios.

An estimate of human performance serves is a useful baseline. However, the task of generating paraphrases that adhere to an extensive set of 40 linguistic attributes is beyond the capabilities of even expert linguists, making human evaluation impractical and potentially unreliable.
Fortunately, we have direct access to the same software tools that precisely compute these linguistic attributes for both the source texts and generated outputs, enabling straightforward and highly accurate automatic evaluation. These deterministic tools calculate exact values for each attribute with consistent reliability, eliminating the subjectivity and variability inherent in human judgments. Our evaluation framework combines BERTScore, which has shown strong correlation with human judgments of semantic similarity in prior work~\citep{Zhang*2020BERTScore:}, with precise attribute measurements that objectively quantify how closely the generated text adheres to target linguistic specifications.

Regarding the performance of our linguistic discriminator, we evaluated its accuracy on predicting 40 different linguistic attributes. The model achieves an average mean square error of 0.52 on the test set, which is significantly lower than the naive baseline of predicting the mean value for each attribute (MSE=1.0). The accuracy varies by attribute category, with some syntactic features (e.g., clause count, T-unit count) being more accurately predicted (average MSE=0.35) than lexical features like sophisticated word usage (average MSE=0.61). This performance gap reflects the inherent complexity of modeling certain linguistic phenomena and represents a limitation of \textsc{LingConv}.

\section*{Ethical Statement}
Controlled text generation needs ethical considerations. There is a fine line between controlled generation and manipulation. Malicious actors may use such a model for the propagation of biased, misleading, or harmful information. We must ensure that the technology is disseminated responsibly, with safeguards in place to prevent malicious usage and unintended consequences. Furthermore, these models allow for generating paraphrases with great diversity that may be undetectable in cases of plagiarism. More sophisticated safeguards around plagiarism, cheating, and theft must be put in place to address this issue.

\section*{Broader Impacts}
The implications of our complexity-controlling paradigm are wide-ranging and significant. By generating more accessible text, this technology extends its reach to individuals with limited literacy proficiency, cognitive impairments, learning disabilities, aphasia, or dementia. Moreover, it allows for personalized communication, functions as a valuable tool for linguistic researchers and natural language processing (NLP) experts, and enhances the pedagogical landscape of second language acquisition by dynamically adapting text complexity to match the learner's skill level. 
In addition, our approach addresses the conversion of text complexity through fine-grained control of linguistic attributes. Text complexity plays a crucial role in text readability, comprehension, and propriety for different readers. 
% The ability to control text generation is not only a technological advancement but also a means to enhance user experience. 
For example, an educational platform that dynamically adjusts the complexity of its content to match the reader's proficiency level can enable better comprehension and engagement. Such personalized learning experiences can potentially revolutionize education by adjusting complexity with respect to the learner's capabilities and accommodating a wider range of learners. Our model can also help content creators to tailor their messaging to their target audience. 
% , amplifying the impact of their communication efforts.

\bibliography{custom,anthology_part1,anthology_part2}

\appendix
\clearpage
% \onecolumn

\begin{figure*}[h!]
\centering
\renewcommand{\arraystretch}{1.2}
\begin{tabularx}{\textwidth}{>{\raggedright\arraybackslash}p{0.18\textwidth} X}
\rowcolor{softblue}
\multicolumn{2}{l}{\textbf{\large Example 1: Lexical brevity}} \\
\textbf{Source} & U.N. inspectors later said the documents were old and irrelevant -- some administrative material , some from a failed and well-known uranium-enrichment program of the 1980s . \\
\rowcolor{softgray}
\textbf{Target} & 2 sophisticated words, 5 total words \\
\textcolor{modelblue}{\textbf{LingConv Gen}} & \textcolor{modelblue}{U.N. inspectors said the documents were old.} \\
\textcolor{modelpurple}{\textbf{QCPG Gen}} & \textcolor{modelpurple}{The documents were old.} \\
\multicolumn{2}{l}{
\begin{tabular}{lccc}
\rowcolor{softgray}
\textbf{Attribute} & \textcolor{targetgray}{\textbf{Target}} & \textcolor{modelblue}{\textbf{LingConv}} & \textcolor{modelpurple}{\textbf{QCPG}} \\
Sophisticated words & 2 & \textcolor{modelblue}{2 (0)} & \textcolor{modelpurple}{0 (\textbf{-2})} \\
Total words & 5 & \textcolor{modelblue}{7 (\textbf{+2})} & \textcolor{modelpurple}{4 (-1)} \\
\end{tabular}
} \\
\textbf{Summary} & LingConv matches the sophistication requirement; QCPG omits it. Both simplify, but only LingConv aligns with the target. \\

\hline
\rowcolor{softblue}
\multicolumn{2}{l}{\textbf{\large Example 2: High complexity}} \\
\textbf{Source} & Named in the complaint were former chief executive officers Paul A. Allaire and G. Richard Thoman and former CFO Barry D. Romeril . \\
\rowcolor{softgray}
\textbf{Target} & 10 sophisticated words, 26 words, 4 complex nominals \\
\textcolor{modelblue}{\textbf{LingConv Gen}} & \textcolor{modelblue}{The executives fined included former Chief Executives Paul A. Allaire and G. Richard Thoman as well as former chief financial officer Barry Romeril, all of the company's board members.} \\
\textcolor{modelpurple}{\textbf{QCPG Gen}} & \textcolor{modelpurple}{The executives fined included former Chief Executives Paul A. Allaire and G. Richard Thoman as well as former Chief Financial Officer Barry Romeril, among others.} \\
\multicolumn{2}{l}{
\begin{tabular}{lccc}
\rowcolor{softgray}
\textbf{Attribute} & \textcolor{targetgray}{\textbf{Target}} & \textcolor{modelblue}{\textbf{LingConv}} & \textcolor{modelpurple}{\textbf{QCPG}} \\
Sophisticated words & 10 & \textcolor{modelblue}{8 (\textbf{-2})} & \textcolor{modelpurple}{9 (-1)} \\
Total words & 26 & \textcolor{modelblue}{30 (\textbf{+4})} & \textcolor{modelpurple}{25 (-1)} \\
Complex nominals & 4 & \textcolor{modelblue}{3 (-1)} & \textcolor{modelpurple}{1 (\textbf{-3})} \\
\end{tabular}
} \\
\textbf{Summary} & Both models generate relevant summaries but cannot fully match the high attribute targets. \\

\hline
\rowcolor{softblue}
\multicolumn{2}{l}{\textbf{\large Example 3: Structural complexity}} \\
\textbf{Source} & The puppy tried to get out of the tub. \\
\rowcolor{softgray}
\textbf{Target} & 20 words, 4 verb phrases, 3 clauses, 3 complex nominals \\
\textcolor{modelblue}{\textbf{LingConv Gen}} & \textcolor{modelblue}{A fluffy puppy is trying to get out of the tube while he is holding a plastic drawer.} \\
\textcolor{modelpurple}{\textbf{QCPG Gen}} & \textcolor{modelpurple}{Even though this puppy is extremely sensitive, it is still trying to get out of the cat's water.} \\
\multicolumn{2}{l}{
\begin{tabular}{lccc}
\rowcolor{softgray}
\textbf{Attribute} & \textcolor{targetgray}{\textbf{Target}} & \textcolor{modelblue}{\textbf{LingConv}} & \textcolor{modelpurple}{\textbf{QCPG}} \\
Total words & 20 & \textcolor{modelblue}{18 (-2)} & \textcolor{modelpurple}{19 (-1)} \\
Verb phrases & 4 & \textcolor{modelblue}{3 (-1)} & \textcolor{modelpurple}{3 (-1)} \\
Clauses & 3 & \textcolor{modelblue}{2 (-1)} & \textcolor{modelpurple}{2 (-1)} \\
Complex nominals & 3 & \textcolor{modelblue}{2 (-1)} & \textcolor{modelpurple}{1 (\textbf{-2})} \\
\end{tabular}
} \\
\textbf{Summary} & Both models fail to reach the structural targets; QCPG's paraphrase also drifts semantically. \\

\end{tabularx}

\caption{\textbf{Qualitative comparison of LingConv and QCPG.} For each attribute, the target, each model's value, and the error magnitude are shown. Large errors are bolded.}
\label{tab:qualitative_examples}
\end{figure*}

\section{Qualitative Analysis}
\label{sec:qualitative}
Figure~\ref{tab:qualitative_examples} presents a qualitative analysis of outputs from different models, highlighting the varying capabilities in linguistic attribute control. This analysis provides insight into the strengths and limitations of our approach compared to existing methods.

\section{List of Linguistic Attributes}\label{app:indices}
\label{sec:index_list}

% \begin{table}[t]
%     \centering
%     \begin{tabular}{ l }
%     \toprule
% \# Unique sophisticated words\\
% \# Unique lexical words \\
% \# Unique sophisticated lexical words\\
% \# Total words\\
% \# Total sophisticated words\\
% Lexical sophistication (unique) \\
% Verb sophistication\\
% Ratio of unique words\\
% Ratio of unique verbs\\
% Ratio of unique adjectives\\
% Ratio of unique adverbs\\
% \# Dependent clauses \\
% \# Clauses \\
% \# T-units \\
% \# Complex T-units \\
% \# Complex nominals \\
%  \# Stop Words\\
%  \# Sentences\\
%  \# Characters\\
%  Average Words Per Sentence\\
%  Average Characters Per Sentence\\
%  Average Characters Per Word\\
%  Average Syllables Per Sentence\\
%  Total Age Of Acquistion Of Words\\
%  \# Named Entities Norp\\
%  \# Named Entities Gpe\\
%  \# Named Entities Law\\
%  \# Named Entities Money\\
%  \# Named Entities Ordinal\\
%  \# Coordinating Conjunctions\\
%  \# Nouns\\
%  \# Numerals\\
%  \# Proper Nouns\\
%  \# Subordinating Conjunctions\\
%  Automated Readability Index\\
%  Reading Time For Average Readers\\
%     \bottomrule
%     \end{tabular}
%     \caption{Linguistic indices used in this paper.}
%     \label{tab:indices}
% \end{table}
\begin{table}[t]
    \centering
    \small
    \begin{tabular}{p{0.52\linewidth} p{0.38\linewidth}}
    \toprule
    \multicolumn{2}{c}{\textbf{Linguistic Attributes}} \\
    \midrule
    Characters per Sentence & Word Count \\[2pt]
    Words per Sentence & Character Count \\[2pt]
    Syllables per Sentence & Sentence Count \\[2pt]
    Characters per Word & Clause Count \\[2pt]
    NORP Entities & T-unit Count \\[2pt]
    GPE Entities & Noun Count \\[2pt]
    Law Entities & Numeral Count \\[2pt]
    Money Entities & Stop Words \\[2pt]
    Ordinal Entities & Proper Nouns \\[2pt]
    Sophisticated Words & Complex T-units \\[2pt]
    Sophisticated Word Count & Complex Nominals \\[2pt]
    Sophisticated Lexical Words & Dependent Clauses \\[2pt]
    Lexical Sophistication & Verb Sophistication \\[2pt]
    Unique Word Ratio & Total Words \\[2pt]
    Unique Verb Ratio & Unique Lexical Words \\[2pt]
    Coordinating Conjunctions & Unique Adverb Ratio \\[2pt]
    Subordinating Conjunctions & Unique Adjective Ratio \\[2pt]
    Age of Acquisition Score & Readability Level \\[2pt]
    \bottomrule
    \end{tabular}
    \caption{Linguistic indices used in this paper.}
    \label{tab:indices}
\end{table}

We use expert-crafted linguistic indices as the control attributes for CPG. Table~\ref{tab:indices} lists all the indices that we use. We select 40 indices, such that there are no duplicates, there is a representative index from each family, there is at least one index from each domain, the index is not too granular as to not be useful, and the selected included indices have utility in text style control.

The linguistic indices employed in our work are derived from off-the-shelf tools that implement linguist-defined rules grounded in psycholinguistics literature. Lexical and surface-level indices, for example, are computed by simply counting word occurrences, ensuring robust and reliable measurements. Syntactic and discourse indices are extracted using part-of-speech (POS) tagging, dependency parsing, and named entity recognition (NER), for which we employ the \texttt{en\_core\_web\_sm} model from spaCy~\citep{spacy}. This model has been reported to achieve 97\% accuracy on POS tagging, 92\% on parsing, and a 0.84 F1-score on NER, verifying the reliability of these derived attributes. Moreover, the deterministic nature of these algorithms guarantees consistent results across experiments. All models and baselines in our study are evaluated using the same input indices and evaluation process, ensuring fair comparison of their linguistic-control capabilities.
For the full descriptions please refer to~\citet{Lu_2010},~\citet{lu2012relationship}, and~\citet{lee-lee-2023-lftk}.

The following is a brief description of a few indices as an example:  
\textbf{Automated Readability Index} is the grade level required for a reader to comprehend the text, from preschool to professor level.
\textbf{Lexical words} are nouns, verbs, adjectives, and adverbs.
\textbf{Sophisticated words} are the unconventional words. We consider the 2000 least frequent words in the American National Corpus as sophisticated. \textbf{GPE Entity} is a geopolitical entity. \textbf{NORP entity} is nationalities or religious or political groups. \textbf{Age of acquisition} is the typical age at which a person learns and begins to use a particular word.
% \textbf{Uber index} is a transformation of TTR.
% \textbf{TTR} is the ratio of unique words in the text.
% \textbf{D-measure} is a modification to TTR that is not biased by sample size.

\section{Comparison of Injection Methods}
\label{app:injection_methods}
Table~\ref{tab:injection_methods} presents a comparison of different injection methods for integrating linguistic attributes into the generation process.
Results are reported on both standard and novel targets test sets in terms of mean squared error for target ($l^t$) and source ($l^s$) linguistic attributes, as well as BERTScore (BERTS).

We experimented with adding linguistic embeddings to all tokens of the decoder input, concatenating to the decoder inputs (equivalent to prompt tuning), concatenation/addition to encoder inputs, concatenating/adding to encoder outputs, and fusing to encoder outputs using a linear layer. In general, decoder injections were better than encoder injections. Decoder first-token-addition was the best-performing overall.

\begin{table*}[ht]
\centering
\small
\begin{tabular}{lccc|ccc}
\toprule
\multirow{2}{*}{Injection Method} & \multicolumn{3}{c|}{} & \multicolumn{3}{c}{Novel Targets} \\
\cmidrule(lr){2-4} \cmidrule(lr){5-7}
 & MSE($l^t$) & MSE($l^s$) & BERTS & MSE($l^t$) & MSE($l^s$) & BERTS \\
\midrule
Encoder Input Concatenation & 0.61 & 0.89 & 94.8 & 4.85 & 9.46 & 86.7 \\
Encoder Input Addition      & 0.62 & 1.08 & 94.3 & 6.31 & 12.75 & 85.2 \\
Decoder Input Concatenation & 16.52 & 18.60 & 85.5 & 28.99 & 38.16 & 82.3 \\
Decoder Input Addition & 0.59 & 0.94 & 94.2 & 10.90 & 15.62 & 85.7 \\
Decoder Input Addition to First Token & 0.58 & 0.91 & 95.1 & 4.32 & 11.55 & 69.0 \\
Layer 1 Addition (all tokens)    & 0.56 & 1.03 & 94.0 & 7.25 & 8.92 & 84.3 \\
Layer 1 Addition (first token)    & 10.01 & 10.20 & 82.0 & 56.63 & 59.79 & 80.1 \\
Layer 6 Addition (first token)    & 12.45 & 12.89 & 83.1 & 62.31 & 65.44 & 81.2 \\
Layer 12 Addition (all tokens)    & 15.67 & 16.02 & 80.5 & 68.92 & 71.35 & 78.9 \\
Layer 12 Addition (first token)    & 16.89 & 17.25 & 79.8 & 71.54 & 73.88 & 77.5 \\

\bottomrule
\end{tabular}
\caption{Comparison of injection methods for linguistic attribute integration. Results include mean squared errors (MSE) for the target ($l^t$) and source ($l^s$) attributes and BERTScore (BERTS), measured on in-distribution (ID) and out-of-distribution (OOD) test sets.}
\label{tab:injection_methods}
\end{table*}

\section{Quality Control Algorithm Details}
\label{app:qc_details}
The quality control (QC) mechanism is designed to optimize the alignment of generated text with target linguistic attributes while preserving semantic equivalence to the source. The QC loop adjusts the text embeddings iteratively using a gradient-based method combined with a line search to minimize attribute errors, following the scientific approach outlined in~\citet{padmakumar2023extrapolative,armijo1966minimization,boyd2004convex}.

The process consists of two key components:
\begin{enumerate}
    \item An iterative refinement process that repeatedly updates the generation until it matches the target attributes or further improvement becomes impossible.
    \item A line search algorithm that finds the optimal control strength for each refinement step while preserving semantic coherence.
\end{enumerate}

Algorithm~\ref{alg:qc} shows this process. Initially, we freeze the parameters of the generation model and set input sentence embeddings as our parameter of interest. The model then generates an initial output $\hat{t}$ (line 4 in Algorithm~\ref{alg:qc}). We use the linguistic attribute predictor component to predict the linguistic attributes of this generation and compute the mean squared error, $l_0$, between the predicted attributes and the target attributes (line 5). The gradient $g$ of this error with respect to the input embeddings provides the direction for updates (line 6).

The adaptive step size is determined through a modified line search algorithm (lines 11--31) that finds the smallest viable step size that improves the output. The resulting generation must satisfy two conditions:
(a) The predicted linguistic attribute error should decrease (i.e., be less than $l_0$)
(b) The semantic equivalence probability should remain above a threshold $\tau$

These conditions ensure both improved attribute control and semantic preservation. The process continues until no viable step size can be found, indicating the generation has reached its optimal state.

\section{Algorithm Background}
This section describes further details on the STE and line search algorithms.

\subsection{Straight-through Gradients}
\label{app:ste}
STE~\citep{bengio2013estimating} is a technique used to propagate gradients through non-differentiable equations in the computational graph, through an estimation of the derivative.
In our case, the decoder produces token logits, which are then transformed into probabilities through softmax. Then, we transform the probabilities into an output sequence using argmax. LP takes as an input the sequence of tokens and not the sequence of logits. However, if we want to propagate the gradient of the loss generated by LP to the main model, we must pass the gradient through the output logits. Thus, we use the following trick to create a pathway in the computational graph from LP's inputs to the logits. First, the output sequence is represented in one-hot encoding rather than a sequence of tokens. Second, we add the logits to the one-hot encoding and subtract a detached (constant) variable equal to the logits. The end result would be equal to the one-hot encoding, but the computational graph now has a path from the logits to LP through the multiplication of the one-hot encoding with LP's text embedding. This means that the gradient propagated to each token of the logits is scaled according to the weights of the text embedding matrix.

\subsection{Line Search}
\label{app:line-search}
Line search~\citep{armijo1966minimization} is a standard numerical optimization algorithm, where at every update step, the step size is chosen dynamically. There are different methods of finding the best step size. They often include trying out many different step sizes, evaluating the resulting parameters, and choosing the step size that results in the lowest loss value. 

Our algorithm is based on backtracking line search, which starts with a large candidate step size, and if it doesn't result in a lower loss than the current, reduce it by a factor of $\gamma$ (often $= 0.5$) and try again. The intuition is that we would like to take the largest step possible that results in an improvement to descend toward the global minimum and potentially avoid local minima. 
However, we would like the opposite; we would like to take the smallest possible step that results in an improvement to not deviate away from the original sentence semantics. Therefore, our algorithm starts from a small step size and grows it by a factor of $\gamma$ at each line search step.

\section{Datasets}
\label{sec:data}
\begin{table}[t]\small
    \centering
    \begin{tabular}{l l l}
         \textbf{Dataset} & \textbf{Full Dataset} & \textbf{Positive Samples} \\
         \toprule
         \textbf{QQP} & 363,846 & 134,378 \\
         \textbf{MRPC} & 3,668 & 2,474 \\
         \textbf{STS-B} & 5,749 & 2,994 \\
         \textbf{Total} & 373,263 & 139,846 \\
    \end{tabular}
    \caption{QQP, MRPC, and STS-B contain samples that are either semantically equivalent or not equivalent. We select from the three datasets samples with the {\it equivalent} label for training and evaluating our model.}
    \label{tab:data_count}
\end{table}

\begin{table*}[t]
    \centering
    \small
    \begin{tabular}{l l llll l}
 & \textbf{Words} & \textbf{Sophisticated Words} & \textbf{Lexical Words} & \textbf{Ratio of Unique Words} & \textbf{Nouns} & \textbf{Readability Index} \\
\toprule
\textbf{Ref} & 12.97 & 4.29 & 7.60 & 9.13\% & 2.16 & 6.62 \\
\textbf{Copy} & 12.98 & 4.29 & 7.61 & 9.25\% & 2.14 & 6.65 \\
\textbf{T5-FT} & 12.83 & 4.22 & 7.49 & 9.18\% & 2.10 & 6.69 \\
\midrule
\textbf{Llama} & 12.04 & 4.55 & 7.25 & 8.29\% & 2.36 & 8.01 \\
\textbf{BOLT} & 10.85 & 3.36 & 6.11 & 8.51\% & 1.83 & 5.47 \\
\textbf{FUDGE} & 11.10 & 3.36 & 6.29 & 7.95\% & 2.00 & 5.09 \\
\textbf{QCPG} & 5.34 & 2.83 & 3.62 & \underline{5.93}\% & \textbf{1.16} & \underline{3.04} \\
\textbf{Lingconv} & \underline{4.37} & \underline{2.38} & \underline{3.04} & \textbf{5.92}\% & 1.27 & 3.36 \\
\hspace{0.6cm}\textbf{+QC} & \textbf{3.21} & \textbf{1.97} & \textbf{2.36} & 6.38\% & \underline{1.23} & \textbf{3.01} \\
    \bottomrule
    \end{tabular}
    \caption{A detailed breakdown of model performance across a selected set of linguistic attributes. performance is reported in mean absolute error (MAE).
    the results are based on novel targets of linguistic attributes.}
    \label{tab:indices_results}
\end{table*}
We combine The Microsoft Research Paraphrase Corpus (MRPC)~\citep{dolan-brockett-2005-automatically}, The Semantic Textual Similarity Benchmark (STS-B)~\citep{cer-etal-2017-semeval}, and The Quora Question Pairs. The three datasets are created for the task of classifying whether the pair of texts are semantically equivalent. Therefore, we only select the positive samples for our model's training and discard the remaining samples. The data distribution is shown in Table~\ref{tab:data_count}. 

The dataset is randomly split into training, validation, and testing sets according to the ratio 80:10:10. The same data is used for training all versions of our approach and baselines. The semantic equivalence and linguistic predictor models are both pre-trained using the same data and splits.

\section{Training Data Preparation}
\label{app:training_data}
We utilize the bidirectional equivalence inherent in paraphrase pairs to enrich our training set with augmented data. First, we augment the data by reversing the order of source and target sentences: $\{t_i, s_i, l_i^t, l_i^s\}$. Second, we augment the data with self-paraphrase pairs: $\{s_i, s_i, l_i^s, l_i^s\}$ and $\{t_i, t_i, l_i^t, l_i^t\}$. This ensures diversity in the types of linguistic conversions that the model can learn, and strengthens the semantic consistency within and across paraphrase pairs, which improves model's understanding and generation capabilities.
We augment 25\% of the training data.

This augmentation strategy significantly increases the diversity of our training data. Starting with our original dataset of approximately 370k samples, our augmentation approach creates a final training set of roughly 840k samples. By including reversed pairs and self-paraphrase pairs, we expose the model to a wider range of attribute transformation patterns. This diversity is crucial for the model to learn flexible linguistic transformations rather than merely memorizing specific source-target attribute pairs.

Importantly, our model learns to map from a dense, normalized 40-dimensional attribute space to generated text. This approach enables generalization to novel attribute combinations not seen during training, as the model learns a continuous function rather than discrete mappings. The inclusion of self-paraphrase pairs where source and target attributes are identical helps the model learn when to preserve aspects of the input, while the reversed pairs teach it bidirectional transformations between different linguistic styles. Together, these augmentation strategies ensure that \textsc{LingConv} can generate diverse outputs tailored to various attribute specifications rather than being constrained to a limited set of transformations.

\section{Baselines}
\label{app:baselines}

\begin{itemize}
    \itemsep0pt
    \itemindent0pt
    
    \item \textbf{Copy}: the output is a copy of the input text.
    
    \item \textbf{Reference}: the output is the ground-truth target paraphrase from the dataset.
    
    \item \textbf{T5-FT}: a standard T5 model that lacks linguistic attribute control capabilities, fine-tuned on the dataset of paraphrase pairs.
    
    \item \textbf{FUDGE}~\citep{yang-klein-2021-fudge}: controlled text generation with future discriminators performs attribute control by weighting the token-prediction logits according to an attribute classifier of the potential continuations.
    
    \item \textbf{QCPG}~\citep{bandel-etal-2022-quality}, quality controlled paraphrase generation is a state-of-the-art model for controlled generation. 
    Target attributes are discretized into tokens, and added as a prefix to the encoder input.
    
    \item \textbf{BOLT}~\citep{liu-etal-2023-bolt}: a decoding-time algorithm for controlled text generation.
    For each test sample, it learns a set of biases by minimizing the losses of an attribute discriminator model and an LM's perplexity.

    \item \textbf{LLama3 (70B)}~\citep{dubey2024llama}: an instruction fine-tuned LLM.
    
\end{itemize}

\section{Experimental Settings}
\label{sec:experiments}
Before adapting all baselines to the \texttt{flan-t5-base} backbone for our comparative experiments, we first replicated their original results using the official code and recommended configurations provided by the respective authors. This ensured faithful reproduction of their approach.

For inference efficiency, our model takes approximately 25 ms/token, compared to FUDGE (112 ms/token), BOLT (114 ms/token), and LLama (162 ms/token), making it significantly more efficient for practical applications. Detailed hyper-parameter settings are provided in Appendix~\ref{sec:experiments}.

We train our model using a single A100 GPU with a batch size of 40, and a learning rate of $1e-3$ Adam optimizer.
% We set $p$ in the weight annealing to be equal to 75\% of the total steps; the linguistic predictor loss has maximum weight only for the last 25\% of training.
We optimize the hyper-parameters of FUDGE and QCPG. In QCPG, optimized batch size $= 8$, learning rate $= 1e-4$, and we train for a large number of epochs $= 20$ to ensure high performance. In FUDGE, we optimize the update factor and the multiplicative factor $\lambda_{FUDGE} = 0.7$. We use the linguistic predictor described in \S~\ref{sec:model} as an attribute classifier for FUDGE, and weigh the logits according to the inverse of the mean squared error of the prediction's linguistic attributes and the target linguistic attributes. Although FUDGE benefits from not having to train or fine-tune the language model, it is extremely slow at inference time due to the demand of evaluating numerous candidates at each generation step. The parameters for the Algorithm~\ref{alg:qc} are: $\eta_0 = 10^{3}, \gamma = 2.25, \tau = 0.95, k = 4$
All models are run with 1 seed. The random seed used for all data processing and models is $0$. When $k > 1$ random seeds are used, such as in section~\ref{app:augmentation}, seeds are from 0 to $k-1$.

The three augmentation settings are trained for 2 epochs, and the best checkpoint is used. We use a learning rate of $1e-3$, batch size of $40$, and linear learning rate scheduling.

Linguistic attributes are quantized using the KBinsDiscretizer\footnote{\url{https://scikit-learn.org/stable/modules/generated/sklearn.preprocessing.KBinsDiscretizer.html}} with the ``kmeans'' clustering strategy.

The per-dataset results in Table~\ref{tab:dataset_results} demonstrate the consistent superiority of LingConv across all three datasets. On QQP (the largest dataset), LingConv reduces attribute error by 39.5\% compared to QCPG, while maintaining comparable BERTScore. Adding quality control further reduces the error by 15.4\%. Similar patterns are observed on STS-B and MRPC, with LingConv+QC achieving the lowest attribute errors of 2.65 and 3.66 respectively, showing that our approach generalizes well across diverse paraphrase collections regardless of domain or size.

\begin{table*}[t]
    \centering
    \small
    \begin{tabular}{l cc cc cc}
    \toprule
    & \multicolumn{2}{c}{\textbf{QQP}} & \multicolumn{2}{c}{\textbf{STS-B}} & \multicolumn{2}{c}{\textbf{MRPC}} \\
    \cmidrule(lr){2-3} \cmidrule(lr){4-5} \cmidrule(lr){6-7}
    \textbf{Model} & \textbf{B-S} & \textbf{MSE($l^t$)} & \textbf{B-S} & \textbf{MSE($l^t$)} & \textbf{B-S} & \textbf{MSE($l^t$)} \\
    \midrule
    \textbf{Ref} & \textbf{100.0} & 10.22 & \textbf{100.0} & 8.63 & \textbf{100.0} & 10.58 \\
    \textbf{Copy} & 94.6 & 10.16 & 94.1 & 8.68 & 94.6 & 10.77 \\
    \textbf{T5-FT} & 95.0 & 10.44 & 94.0 & 8.64 & 93.5 & 10.59 \\
    \midrule
    \textbf{Llama} & 91.1 & 7.98 & 90.7 & 7.64 & 90.8 & 11.25 \\
    \textbf{BOLT} & 90.5 & 7.92 & 89.5 & 7.04 & 87.8 & 7.76 \\
    \textbf{FUDGE} & 92.6 & 6.79 & 91.1 & 7.07 & 89.5 & 8.04 \\
    \textbf{QCPG} & 90.4 & 6.12 & 90.6 & 4.56 & 89.9 & 6.56 \\
    \textbf{Lingconv} & 90.9 & 3.70 & 91.3 & 3.55 & 90.5 & 4.57 \\
    \hspace{0.6cm}\textbf{+QC} & 90.8 & \textbf{3.13} & 90.8 & \textbf{2.65} & 89.7 & \textbf{3.66} \\
    \bottomrule
    \end{tabular}
    \caption{Performance breakdown by dataset on the Novel Target Challenge. B-S is BERTScore and MSE($l^t$) is mean squared error of target attributes (lower is better). Results show LingConv consistently outperforms baselines across all datasets, with the quality control mechanism providing further improvements.}
    \label{tab:dataset_results}
\end{table*}

\begin{table}[t]
    \centering
    \small
\begin{tabular}{lllll}
\textbf{Model} & \textbf{Lexical} & \textbf{Syntactic} & \textbf{Discourse}  & \textbf{Macro-}\\
& & & & $\textbf{MSE} (\pmb{l^t})$ \\
\toprule
\textbf{Ref} & 12.62 & 8.89 & 5.91 & 9.14 \\
\textbf{Copy} & 12.66 & 8.87 & 6.19 & 9.24 \\
\textbf{T5-FT} & 12.73 & 8.82 & 6.16 & 9.24 \\
\midrule
\textbf{Llama} & 10.88 & 8.37 & 5.56 & 8.27 \\
\textbf{BOLT} & 9.36 & 7.23 & \underline{3.21} & 6.60 \\
\textbf{FUDGE} & 9.54 & 6.83 & \textbf{2.34} & 6.23 \\
\textbf{QCPG} & 7.64 & 4.30 & 5.46 & 5.80 \\
\textbf{Lingconv} & \underline{4.25} & \underline{3.08} & 4.70 & \underline{4.01} \\
\hspace{0.6cm}\textbf{+QC} & \textbf{3.51} & \textbf{2.31} & 3.62 & \textbf{3.15} \\
\bottomrule
\end{tabular}
    \caption{A detailed breakdown of model performance (MSE) across distinct groups of linguistic attributes. 
    Each group represents specific linguistic attributes that contribute to the overall complexity and structure of the generated text. }
    \label{tab:types_results}
\end{table}

\section{Analysis of Linguistic Attributes}
\label{app:analysis}
We analyze the performance of models across different groups of linguistic attributes to understand their strengths and weaknesses, and the inherent difficulty in controlling different types of attributes.
We group the linguistic attributes into several types according to the categorizations in~\citep{Lu_2010, lu2012relationship, lee-lee-2023-lftk}. The attribute types are lexical, syntactic, and discourse features.
We analyze MSE values for each model across standard and novel target scenarios, revealing the following insights:

\subsection{Controlling Discourse Proves Most Challenging}
\label{sec:types_analysis}
Table~\ref{tab:types_results} shows the error rate of each approach in controlling different linguistic 
%index 
attribute groups. 
Despite having the lowest average error across models, discourse attributes show the smallest reduction in error by \textsc{LingConv} compared to T5-FT, at 41\%. 
This suggests that discourse attributes are the most challenging to control.
In contrast, lexical attributes have the highest average baseline error, and \textsc{LingConv} achieves the most significant reduction in this error, at 74\%.
Syntactic attributes appear to be the easiest to control, with the error rate dropping from 8.82 to 2.31, a 73\% reduction, the lowest among all groups.
We note that FUDGE achieves the lowest error in discourse attributes. 
% In our study, 
This is because many of these attributes are represented by the presence and density of particular named entities. The generation of FUDGE is driven by the next word that minimizes the MSE. Therefore, it can generate the singular named entities that significantly reduce the error. However, this is not an optimal strategy for syntactic structures that require several iterations of planning and building, as evidenced by the high error rate of FUDGE on syntactic attributes.\looseness-1
% On the other hand, our approach works to produce generations that holistically follow the target attributes.

\paragraph{Quality Control Boosts Adherence across Linguistic Attributes}
The quality control algorithm reduces the error rates of \textsc{LingConv} across all types of attributes. 
The largest improvement of 25\% is in syntactic attributes. The algorithm of iterative refinement of a source sentence is particularly suited to the task of iteratively adding and deleting selected entities, and matching the required target more closely. 
The second largest improvement is in lexical attributes at 23\%, the algorithm can iteratively add and delete selected words, matching the desired lexicon and minimizing the error in lexical attributes.
Finally, discourse features often require a complete restructuring of the sentence, which is the most difficult. However, quality control achieves a 17\% reduction in error.

% surface feats -> syntax
% The attributes in this group describe the number of words in a sentence and the ease of readability. This is because a paraphrase of a sentence typically comprises a similar sentence structure, so it is easy to match those target attributes. On the other hand, discourse attributes are the hardest to control. This group of attributes describes the insertion and deletion of particular types of entities into the sentence. This task is challenging, as the generated paraphrase has high flexibility in the inclusion and exclusion of entities that do not affect the main ideas of the text.

% Lexico-semantic -> lexical
% The easiest group is lexico-semantic. This group describes words in terms of variation and age-of-acquisition. It appears that it is easy for auto-regressive text conversion models to generate text with repeated usage of the same words or with new words. The most difficult group to control is lexical features. This group includes attributes that describe the word sophistication and word type variation. We look more closely into the attributes of this group and find that the increased error is the result of three attributes: the unique number of words in the first $k$ tokens, in a random sub-sequence of length $k$, and in a random selection of $k$ tokens. These constraints cannot be arbitrarily matched while preserving the semantics of the source sentence, therefore their performance drops in the novel target scenario.

To further verify, we apply the quality control mechanism to T5-FT, instead of \textsc{LingConv}. T5-FT plus quality control has a 0.90 MSE($l^t$) in the standard case and 9.20 in novel target case. In both scenarios, the model improved over the vanilla T5. However, it is evident from this results that quality control alone is not sufficient for attribute control, and the architecture of \textsc{LingConv} is essential.

Our model achieves a 34\% error reduction over LLama in attribute control.
While large models like LLama-70B excel at general-purpose generation, our results show they struggle with precise attribute control (MSE($l^t$) of 8.90 vs LingConv's 3.69 in novel target scenarios).
The quality control mechanism provides an additional 14\% error reduction, demonstrating the value of having full access to model gradients. These results suggest that the black-box nature of prompt-based approaches limits their ability to achieve exact attribute matching.

\paragraph{Linguistic Predictor Performance}
The final MSE loss %in 
of the pre-trained linguistic predictor (LP) is $0.16$ on our test set, indicating that the model's results have been achieved despite using imperfect linguistic predictor. This could potentially compound errors in the refined outputs generated during inference time with quality control mechanism. 
% To address this concern, we measure the correlation between the predicted MSE by the LP and the real MSE by the original linguistic attribute extractor tool on the refined outputs. We find that the correlation between the two is $0.8$, which is sufficiently high for us to reliably utilize the LP. 
We further report the error of the linguistic discriminator over different types of attributes in Table~\ref{tab:ling-disc-loss}. We find that the error rates are lowest for lexical attributes, moderately higher for syntactic attributes, and highest for discourse attributes. This finding is consistent with the literature on linguistic attributes~\citep{pallotti2019approach,rafatbakhsh2023predicting}. 

\subsection{Handling of Contradictory Attributes}
In this paper, we assumed all input control vectors are linguistically valid. To extend our work to handle potentially contradictory attributes, a clear definition of such conflicts is needed. Contradictions can arise from structurally incompatible requirements (e.g., a control vector specifying both ``no verbs'' and ``three verb phrases'') or from attributes that are strongly negatively correlated (e.g., requesting a ``higher number of clauses'' alongside a ``lower average sentence length''). In such cases, we hypothesize that the model would prioritize the more commonly observed attribute pattern from its training data, effectively ignoring the outlier request. For instance, it would likely ignore a ``no verbs'' constraint in favor of producing a grammatically plausible sentence with verb phrases. Analyzing model behavior with conflicting targets is a valuable direction for future work, and could reveal insights into the model's implicit linguistic biases and trade-off strategies.

\section{Attribute-specific Performance}
Table~\ref{tab:indices_results} shows the error rate of each approach with respect to individual attributes. The errors are reported in mean absolute error (MAE).

LingConv achieves the least error in 5 out of 6 of the listed indices.
LLama shows the worst performance compared to CPG methods.
Compared to the T5-FT baseline, BOLT and FUDGE only slightly improve the error. QCPG is the best-performing baseline after LingConv.
Notably, QCPG shows the smallest error in controlling the number of nouns in a sentence. Moreover, QCPG controls the readability index of the generation with an MAE of 3 and the ratio of unique words in a sentence with an error of 6\%. For both of these indices, LingConv still achieves the smallest error. 

LingConv controls the number of words up to an error of 3 words, which is the best among all baselines. LingConv also significantly improves upon the control of word sophistication in the sentence, with an MAE of 2 words. Finally, LingConv can control the reading level of a sentence from Kindergarten (1) to Professor (14) level with an MAE of 3, which is non-trivial given that non-control baselines have an MAE of 6 levels, and LLama has an MAE of 8 levels.

\section{Imputation of Missing Values}
\label{app:imputation_details}

Missing linguistic attribute values are imputed using the Multiple Imputation by Chained Equations (MICE) algorithm~\citep{azur2011multiple}. For each missing attribute, a regression model is fitted using the other observed variables, and missing values are imputed based on this model. This process is repeated for 1000 iterations for each variable with missing data, forming a chain of equations that leads to iterative refinement. We use a Ridge Regression~\citep{golub1999tikhonov} linear model as the estimator with $\alpha=1000$ to ensure robust predictions.

The regression models are fitted using a training set consisting of ground-truth linguistic attribute vectors from the training data of \textsc{LingConv}. Before the initial iteration of MICE, missing values are initialized using the mean value for each attribute, allowing prediction of a missing attribute as a function of all other attributes.

MICE leverages the relationships among variables to handle missing data. In the context of linguistic attributes, there are fixed relations between many of the attributes (e.g., any lexical count cannot be larger than the total number of words, and the number of clauses cannot be larger than the number of sentences in the text). By using ridge regression models within the MICE framework, these relationships are preserved while providing regularized predictions that avoid overfitting. This imputation mechanism enables users to specify only a few attributes of interest rather than all 40 attributes.

\subsection{Imputation Experiment Results}
\label{app:imputation}
We evaluate the performance of the MICE imputation method. We find that it leads to 0.02, 0.05, 0.06, 0.11 mean squared error, for imputing 20\%, 40\%, 60\%, and 80\% of the attributes, respectively, in the linguistic attributes that are imputed.

\section{Distributions of Augmentation Attributes}
\label{app:aug-data}
\begin{figure}[htb]
    \centering
    \includegraphics[width=\linewidth]{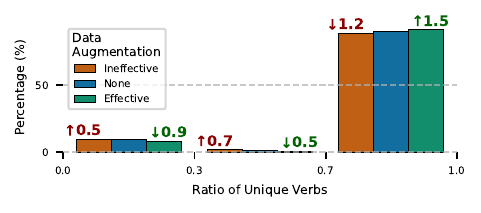}
    \vspace{-5pt}
    \caption{For CoLA (Limited), effective augmentation is associated with increased percentage of sentences with ratio of unique verbs $> 0.7$.}
    \label{fig:aug-cola-l}
\end{figure}

\begin{figure}[htb]
    \centering
    \begin{subfigure}[t]{\linewidth}
        \centering
        \includegraphics[width=\linewidth]{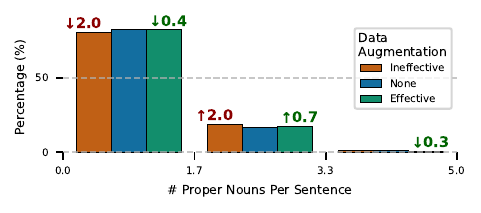}
        \caption{Increase the percentage of sentences with a smaller number of proper nouns.}
    \end{subfigure}%
     
    \begin{subfigure}[t]{\linewidth}
        \centering
        \includegraphics[width=\linewidth]{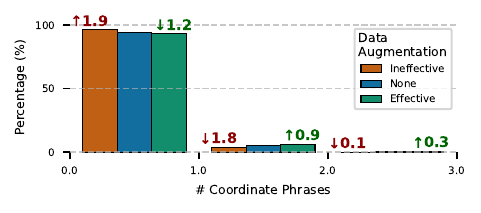}
        \caption{Increase the number of sentences with more than 1 corodinate phrase.}
    \end{subfigure}
    \caption{For CoLA (Full), we bias two attributes.}
    \label{fig:aug-cola-f}
\end{figure}

\begin{figure}[htb]
    \centering
    \begin{subfigure}[t]{\linewidth}
        \centering
        \includegraphics[width=\linewidth]{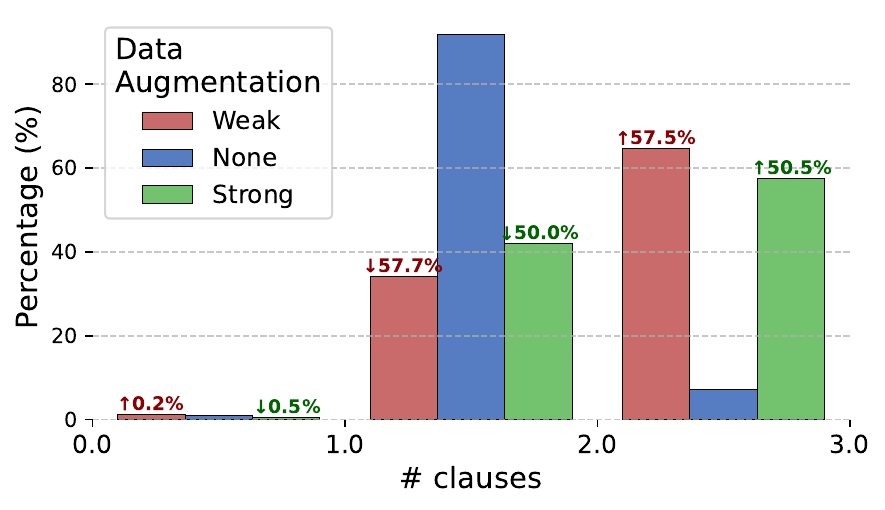}
        \caption{Increase the percentage of sentences with more than 3 clauses.}
    \end{subfigure}%
    
    \begin{subfigure}[t]{\linewidth}
        \centering
        \includegraphics[width=\linewidth]{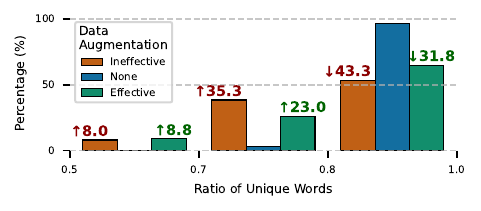}
        \caption{Decrease the percentage of sentences with TTR $> 0.8$.}
    \end{subfigure}
    \caption{For RTE (Full), we bias two attributes.}
    % \caption{For RTE (Full), we increase the percentage of sentences with a high ratio of unique words (Type-token Ratio).}
    \label{fig:aug-rte-f}
\end{figure}

\begin{figure}[htb]
    \centering
    \includegraphics[width=\linewidth]{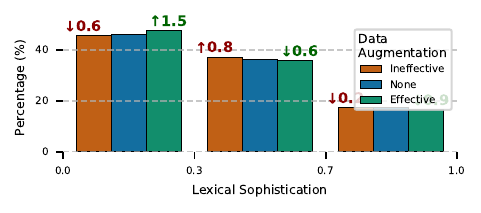}
    \vspace{-5pt}
    \caption{For SST-2 (Limited), decrease the ratio of sophisticated words.}
    \label{fig:aug-sst2-l}
\end{figure}

\begin{figure}[htb]
    \centering
    \begin{subfigure}[t]{\linewidth}
        \centering
        \includegraphics[width=\linewidth]{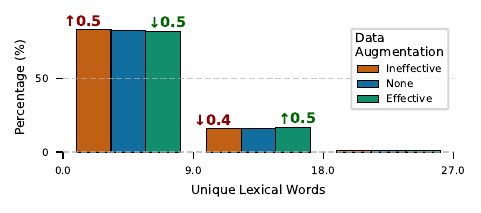}
        \caption{Increase number of unique lexical words.}
    \end{subfigure}%
    
    \begin{subfigure}[t]{\linewidth}
        \centering
        \includegraphics[width=\linewidth]{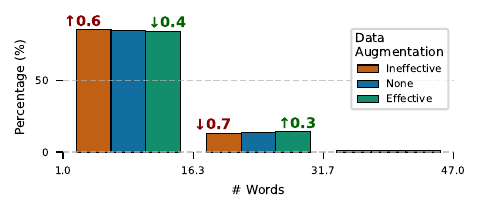}
        \caption{Increased average sentence length.}
    \end{subfigure}
    
    \begin{subfigure}[t]{\linewidth}
        \centering
        \includegraphics[width=\linewidth]{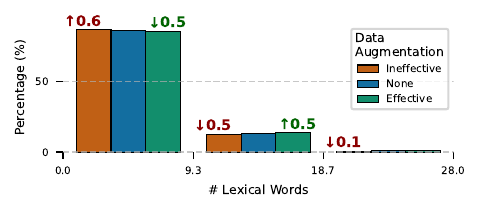}
        \caption{Increase sentences with \# Lexical Words $> 9$}
    \end{subfigure}
    \caption{For SST-2 (Full), we bias the number of lexical words, total words, and unique lexical words.}
    \label{fig:aug-sst2-f}
\end{figure}

Figures~\ref{fig:aug-cola-l}-\ref{fig:aug-sst2-f} show the distributions of the biased attributes in the strong and weak sets of target linguistic variables.

Figure~\ref{fig:aug-cola-l} shows that for the CoLA (Limited) dataset, effective augmentation is correlated with an increased percentage of sentences where the ratio of unique verbs exceeds 0.7. This suggests that sentences with a higher diversity of verbs contribute to more effective augmentation, likely by enhancing the semantic richness of the generated data.

Figure~\ref{fig:aug-cola-f} presents results for the CoLA (Full) dataset with two distinct attribute biases. On the left, we see that increasing the percentage of sentences with fewer proper nouns is associated with effective augmentation. This indicates that simpler sentences with fewer proper nouns may improve performance. On the right, the data shows that increasing the number of sentences containing more than one coordinate phrase also leads to effective augmentation. This suggests that complex sentence structures with multiple coordinate phrases contribute positively to augmentation effectiveness.

Figure~\ref{fig:aug-rte-f} details biases applied to the RTE (Full) dataset. The left subplot indicates that effective augmentation is linked to a higher percentage of sentences with more than three clauses. This suggests that sentences with more complex structures are beneficial for augmentation. Conversely, the right subplot shows that decreasing the percentage of sentences with a Type-Token Ratio (TTR) greater than 0.8 is associated with effective augmentation. This implies that sentences with a lower TTR, reflecting less lexical variety, can also enhance augmentation effectiveness.

Figure~\ref{fig:aug-sst2-l} demonstrates the impact of reducing the ratio of sophisticated words in the SST-2 (Limited) dataset. Effective augmentation is associated with a decrease in sophisticated words, suggesting that simpler vocabulary contributes to better augmentation outcomes in this dataset.

Figure~\ref{fig:aug-sst2-f} provides a detailed view of biased attributes for the SST-2 (Full) dataset. The top-left subplot shows that increasing the number of unique lexical words leads to effective augmentation. The top-right subplot reveals that increasing the average sentence length is also beneficial. Additionally, the bottom subplot indicates that a higher number of sentences with more than nine lexical words contributes to effective augmentation. These results suggest that a richer vocabulary and longer sentences improve augmentation effectiveness.

These figures collectively illustrate how manipulating various linguistic attributes influences the effectiveness of data augmentation, highlighting specific features that can be optimized to enhance performance across different datasets.

\begin{figure}[htb]
    \centering
    \includegraphics[width=\linewidth ]{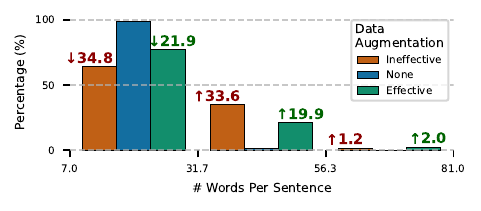}
    \vspace{-15pt}
    \caption{Attribute distributions for effective vs. ineffective augmentation on the RTE (Limited) dataset. Effective augmentation has a greater percentage of shorter sentences.}
    \label{fig:aug-rte-l}
    \vspace{-10pt}
\end{figure}


\end{document}